\documentclass[10 pt, conference]{ieeeconf}  

\IEEEoverridecommandlockouts                              

\usepackage{amssymb,url,amsmath,bm}
\usepackage{algorithm}
\usepackage{graphicx}
\usepackage{epstopdf}
\usepackage{textcomp}
\usepackage{color}
\usepackage{ifpdf}
\usepackage{url}
\usepackage{enumerate}
\usepackage{array}
\usepackage{flushend}
\usepackage{multirow,booktabs}
\usepackage[pdfstartview=]{hyperref}
\usepackage[export]{adjustbox} 
\usepackage{subfig}
\usepackage{mdwlist}
\usepackage[utf8]{inputenc}
\usepackage{todonotes}
\usepackage{hyperref}

\usepackage{cite}

\title{\LARGE \bf
Multisensory Learning Framework for Robot Drumming}

\author{Andrey Barsky*, Claudio Zito*, Hiroki Mori**, Tetsuya Ogata** and Jeremy L. Wyatt*
\thanks{* IRLab, School of Computer Science, University of Birmingham, United Kingdom,
 {\tt\small \{C.Zito\}@cs.bham.ac.uk}}%
 \thanks{** Department of Intermedia Art and Science, Waseda University, Japan}
}

\begin{document}

\maketitle
\thispagestyle{empty}
\pagestyle{empty}

\begin{abstract}

The hype about sensorimotor learning is currently reaching high fever, thanks to the latest advancement in deep learning. In this paper, we present an open-source framework for collecting large-scale, time-synchronised synthetic data from highly disparate sensory modalities, such as audio, video, and proprioception, for learning robot manipulation tasks. We demonstrate the learning of non-linear sensorimotor mappings for a humanoid drumming robot that generates novel motion sequences from desired audio data using cross-modal correspondences. We evaluate our system through the quality of its cross-modal retrieval, for generating suitable motion sequences to match desired unseen audio or video sequences.     

\end{abstract}


\section{INTRODUCTION}\label{sec:introduction}

Sensorimotor learning enables a robot to determine correspondence between sensory modalities (e.g. vision and/or audio) and motor actions~\cite{bib:axenie_2016, bib:levine_2016}. Using deep learning, we can extrapolate multimodal features that represent correspondences between input modalities that are more informative,  facilitating the subsequent learning of desired motor actions. Moreover, as modalities are not always reliable, when one modality becomes corrupted the system can extrapolate the missing information from the other inputs~\cite{bib:ngiam_2011,bib:noda_2014}.

Previous work on multisensory integration in robotics (e.g., solving audio-visual~\cite{bib:axenie_2016} or visuo-tactile correspondence~\cite{bib:levine_2016}) required large-scale data collection on real robot platforms. However there are advantages to a low-cost platform where effective pilot data can be collected in simulation. We present an open-source framework for quick and efficient simulation and recording of synchronised sensory data across multiple modalities. 

We focus on the application of autonomous robot drumming. This task requires an agent to learn the nonlinear multisensory mappings between vision, audio and motor modalities; in particular, the sensorimotor correspondence reflects a kind of causal relationship, i.e.: what sensory percept would result from a given motor movement, or what motor movement would be required to reach a given sensory percept. The experimental results show that drumming motions can be predicted for previously unseen video and audio sequences.


\section{OUR FRAMEWORK}\label{sec:approach}

In this section, we will present i) our open-source framework for generating large-scale, time-synchronised data, and ii) our sensory integration network for sensorimotor learning.    

\subsection{Simulated data collection}

\begin{figure}[t]
\centering
\includegraphics[scale=0.36]{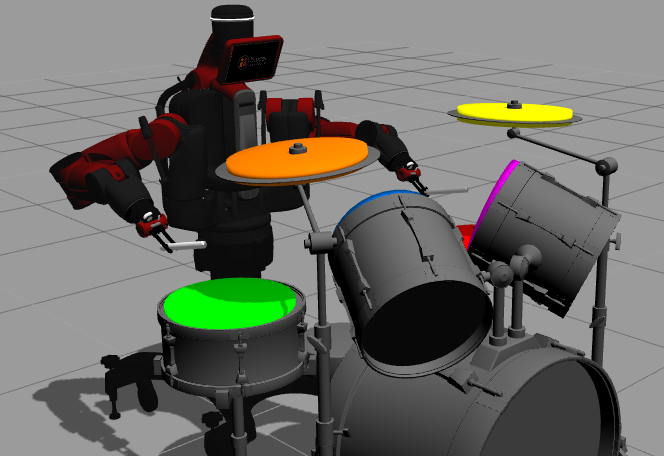}
\includegraphics[scale=0.48]{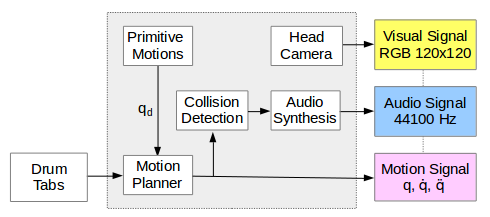}
\caption{\textit{Top:} Simulation setup for drumming task in Gazebo. The coloured surfaces represent target regions that generate audio.
\textit{Bottom:} Schematic of our framework. Input is given as a drum tab, or desired beats for each element of the drumkit. Motion primitives (i.e., joint angles $q_d$) are demonstrated off-line for each element. Our motion planner computes angles, $q$, velocities, $\dot{q}$, and accelerations, $\ddot{q}$, to synchronise the motion primitives to the desired beat times. When a collision is detected, the associated audio signal is generated. Images are captured from the (virtual) robot's head camera. All three sensory signals are captured synchronously and collected into a single multimodal record.}
\label{fig:drummer}
\end{figure}

Our framework is built using the Gazebo software for robot simulation, and we employ the Baxter Research Robot for its thorough SDK and Gazebo integration. The system is integrated into the Robot Operating System (ROS) architecture as a collection of nodes for motion planning, collision detection and audio synthesis. See Figure~\ref{fig:drummer}.

\begin{figure*}[t]
\centering
\includegraphics[scale=0.53]{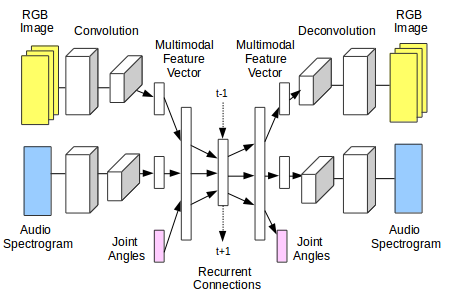}
\includegraphics[scale=0.4]{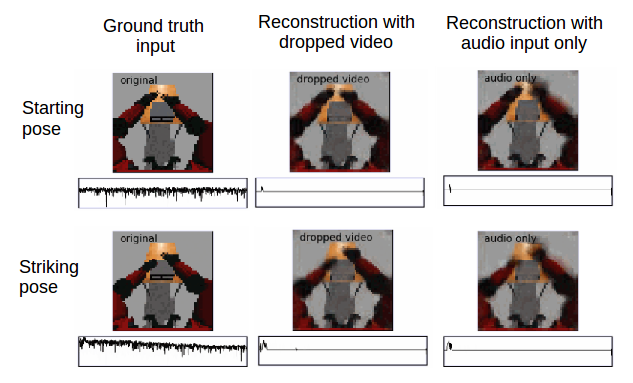}
\caption{\textit{Left:} Multimodal integration network schematic. Image and audio streams at each time step are separately encoded to extract unimodal features. These features are concatenated with each other and with the joint angles into a multimodal feature vector. The multimodal features are used as input to a contractive autoencoder with recurrent LSTM connections in its central layer. This autoencoder reconstructs the multimodal feature vector, which is then decoded to produce separate reconstructions of the three sensory signals. Modalities on the left are randomly dropped out during training. The reconstruction loss is the mean of  mean square errors across the three sensory inputs and their reconstructions, weighting each modality equally.
\textit{Right:} Network reconstructions with dropped out input modalities. Shown are images and audio spectrograms for two frames of a trial with only a snare drum, where we demonstrated a tap on the drum with the left arm. Due to MSE loss, the reconstructed spectrograms average out the white noise of the ground truth input, but retain the signal of the drum beat. The visual reconstructions are qualitatively similar to the ground truth input and follow its movement across time. }
\label{fig:network}
\end{figure*}

We demonstrate to the robot a set of motion primitives to strike the elements of the drumkit. The primitives are represented as linear motions composed of three waypoints: i) a start pose, ii) a contact pose, and iii) an end pose, which is equal to i). In our test scenario, their velocities and accelerations are maintained to produce a constant volume. Movement between primitives is handled by a motion planner to produce a smooth transition from element to element. 

Desired beat times are provided as input to the system, as a series of timestamps and corresponding drum targets (i.e., a drum tablature). For each motion executed, we record the collisions between the robot's end effectors and the drum surfaces, and use these to synthesise audio. 




\subsection{Sensory integration network}

Our sensory integration network follows~\cite{bib:noda_2014} in combining learned representations from individual unimodal convolutional neural networks into a multimodal feature vector, then training an LSTM network to reconstruct these multimodal features (see Figure~\ref{fig:network}, left). To ensure that the representations are being fused, and not simply processed independently, during training we randomly drop out some modalities but still require reconstruction of the full sensory data, following~\cite{bib:ngiam_2011}. 

This is what allows for modality invariance: by forcing the multimodal feature vector to hold the same high-level representations for movements regardless of the sensory modality they are perceived in, we achieve a generalisable mapping from any modality to any other, given sufficient information. 


\section{Discussion \& Future Work} 
\label{sec:experiments}

Preliminary analysis of pilot data shows that the proposed sensory integration system can learn cross-modal retrieval through all combinations of modalities. In particular, motion sequences can successfully be generated from desired audio sequences, meaning that we can predict appropriate drumming motions from given audio context (see Figure~\ref{fig:network}, right). The audio reconstructions retain the drum sound signal and discard the white noise, while the visual reconstructions correctly approximate the movements in the ground truth. 
Motion reconstructions from audio are noisy because of higher-order dependencies in the audio-motor correspondence---the robot must start moving \textit{before} it hears a sound. Future work will focus on improving this reconstruction using a bidirectional RNN framework (to capture these higher-order dependencies).
We will also focus on integrating tactile feedback as a fourth modality, which has greater application for manipulation in other domains.


\section{Repository} 
\label{sec:repository}

Our open-source framework is available under GNU General Public License (GPL) at the following link:\\
\scriptsize{\url{https://github.com/andreybarsky/sensorimotor_drumming}}

%



\bibliographystyle{plain}
\bibliography{iros2018}

\end{document}